\documentclass{article}

\usepackage{microtype}
\usepackage{graphicx}
\usepackage{hyperref}

\usepackage[accepted]{icml2020}

\usepackage{array, boldline, makecell, booktabs}
\newcolumntype{M}[1]{>{\centering\arraybackslash}m{#1}}
\usepackage{subcaption}
\usepackage{mwe}
\usepackage{amssymb}        % blackboard math symbols
\usepackage{amsmath} 
\usepackage{xfrac}
\usepackage{nccmath}
\usepackage{mathtools}
\DeclareMathOperator{\E}{\mathbb{E}}
\icmltitlerunning{Not Your Grandfather's Test Set: Reducing Labeling Effort for Testing}

\begin{document}

\twocolumn[

\icmltitle{Not Your Grandfather's Test Set:  \\
Reducing Labeling Effort for Testing}

\icmlsetsymbol{equal}{*}

\begin{icmlauthorlist}
\icmlauthor{Begum Taskazan}{neu}
\icmlauthor{Jiri Navratil}{ibm}
\icmlauthor{Matthew Arnold}{ibm}
\icmlauthor{Anupama Murthi}{ibm}
\icmlauthor{Ganesh Venkataraman}{ibm}
\icmlauthor{Benjamin Elder}{ibm}
\end{icmlauthorlist}

\icmlaffiliation{neu}{Department of ECE, Northeastern University, Boston, US}
\icmlaffiliation{ibm}{IBM Research AI, Yorktown Heights, US}

\icmlcorrespondingauthor{Begum Taskazan}{taskazan.b@northeastern.edu}
\icmlcorrespondingauthor{Jiri Navratil}{jiri@us.ibm.com}

% You may provide any keywords that you
% find helpful for describing your paper; these are used to populate
% the "keywords" metadata in the PDF but will not be shown in the document
\icmlkeywords{Machine Learning, ICML}

\vskip 0.3in
]

% this must go after the closing bracket ] following \twocolumn[ ...

% This command actually creates the footnote in the first column
% listing the affiliations and the copyright notice.
% The command takes one argument, which is text to display at the start of the footnote.
% The \icmlEqualContribution command is standard text for equal contribution.
% Remove it (just {}) if you do not need this facility.

\printAffiliationsAndNotice{}  % leave blank if no need to mention equal contribution
% \printAffiliationsAndNotice{\icmlEqualContribution} % otherwise use the standard text.

\begin{abstract}
Building and maintaining high-quality test sets remains a laborious and expensive task.  As a result, test sets in the real world are often not properly kept up to date and drift from the production traffic they are supposed to represent.  The frequency and severity of this drift raises serious concerns over the value of manually labeled test sets in the QA process.
This paper proposes a simple but effective technique that drastically reduces the effort needed to construct and maintain a high-quality test set (reducing labeling effort by 80-100\% across a range of practical scenarios). This result encourages a fundamental rethinking of the testing process by both practitioners, who can use these techniques immediately to improve their testing, and researchers who can help address many of the open questions raised by this new approach.
\end{abstract}

\section{Introduction}

Testing is a critical part of the quality assurance process for models used in real-world applications and reducing the time spent on model development.
Traditional testing takes a model and a test set as input, and outputs quality metrics such as accuracy, precision, or recall.
A critical assumption in such a process is that the test set is uniformly and randomly sampled from production traffic. If this assumption is violated the test set results can no longer be used as an indicator of how the model will perform in the production environment.
Unfortunately, production data continually changes over time.
A test set that is perfectly representative today may be hopelessly out of date next week if production data shifts. 
One solution to this problem is to continually refresh the test set by sampling new data from production.  Taken to the extreme, the entire test set could be resampled on a regular basis.
However, labeling data is an expensive and laborious process.
There is typically a strong desire to minimize the amount of labeling that occurs, and when labeling does occur, priority is often given to expanding model training data rather than test sets \cite{settles2009active, bachman2017learning, konyushkova2017learning}.
As a result, test sets often drift out of date and the metrics they produce have little relevance to what is happening in production.
This paper addresses the above problems by proposing a new formulation of testing that drastically reduces the labeling effort required to maintain a high quality test set. It does so by breaking the assumption that the test set distribution needs to match that of the production data.
In our approach,
performance metrics are produced via a {\it performance predictor} which takes as input not only a model and test set, but also a batch of unlabeled production data.   
By having access to both the test set and production data, a performance predictor can observe any distributional differences between them and use this information to compute more accurate predictive metrics.
% This technique is extremely simple but significantly reduces the effort needed to maintain a high-quality test set.
Our experimental results show that this technique can reduce the labeling effort needed to achieve a given test set error by 80\%-100\% over a range of datasets and drift scenarios.  

The main contributions of this paper are as follows: (1) we demonstrate the effectiveness of performance prediction and test set resampling in approximating the model accuracy on an unlabeled dataset, (2) we evaluate several variants of test set update strategies, and (3) draw the attention in labeling effort reduction research from the training stage of learning models to the testing stage, considering the practical scenarios where retraining of the model is expensive and possibly unnecessary. Results are presented on multiple datasets spanning a variety of classification tasks.

\subsection{Problem Formulation}
\label{Sec:Problem Formulation}
Our technique assumes to have a trained base model $\mathcal{M}$, a batch of $N$ unlabeled data samples $X=\{x_1, ..., x_N\}$, and constant labeling cost per individual data point. 
Given a budget to label up to $k<N$ points from $X$, let  $S$, $|S|=k$, be an index subset identifying elements of $X$ that have been labeled, with $Y_S$ the set of the corresponding labels. 
We seek to generate an accuracy estimate $\widehat{Acc}$ using only the labels $Y_S$ such that the following quantity is minimized:
\begin{equation}
\label{Eq:AccDiff}
|\E[\widehat{Acc}(\mathcal{M}|Y_S,X)]-\E[Acc(\mathcal{M}|Y_{all},X)]| 
\end{equation}
Above, $Y_{all}$ denotes a set of labels for the entire dataset $X$ and $\E$ denotes the expectation. While $Acc(\mathcal{M})$ refers to the conventional accuracy calculation for a model $\mathcal{M}$, $\widehat{Acc}(\mathcal{M})$ stands for any function generating an estimate of such a metric, in particular, including output of predictive models. 
Further expanding the above scenario: suppose there is an infinite sequence of batches $X_1, X_2, ...$. The joint distribution of samples and labels, $P(X,Y)$, may change from batch to batch and retraining is expensive. Suppose we maintain a working test set $T$ as a result of processing past batches $X_1, ..., X_{t-1}\,\,$. For $X_t$, the following questions are relevant: is there an optimum strategy to (a) {\em add} (label) and, (b) {\em remove} previously labeled points from the test set $T$, s.t. Eq. (\ref{Eq:AccDiff}) is minimized on $X_t$?
While the above formulation asks for an optimum strategy, such a strategy may only exist under certain assumptions, e.g., for certain model types and data constraints. In general, the above problem is underdetermined and we defer a theoretical analysis to future work. In this paper, we tackle a relaxed version of the problem, namely generating accuracy estimates that require fewer--albeit not minimum--labeled points compared to their conventional alternative. We show empirically that a better approximation of $Acc(\mathcal{M}|Y_{all},X)$ can be found by using these three methods: performance prediction, uncertainty prioritization, and resampling, while labeling only $k$ samples to stay within budget. 

\subsection{Related Work}
Model testing in machine learning (ML) is generally defined as the \textit{oracle problem} in that the ground truth for the model output is not available. This definition is valid for both supervised and unsupervised learning  \citep{xie2018mettle}. Several studies use adversarial samples to reveal the model's failure cases, and the model is retrained based on those ``adversarial" samples. To enhance this formulation, several prioritization methods are suggested to find ``best" samples first \citep{zhang2019noise,byun2019input}. Instead of detecting failure cases, ML testing can also be formulated as an accuracy  prediction problem \citep{li2019boosting}. The advantage of this approach is in taking the operational context 
into account, unlike with the adversarial example based methods. A more extensive survey on machine learning testing workflow, its importance and its components can be found in \cite{zhang2019machine}. 

 In this paper, we also consider testing as an accuracy (or performance) prediction problem. However, we tackle the problem considering the full model life-cycle. In the work of \cite{li2019boosting}, a trained model and an unlabeled dataset (operational data) is leveraged to determine which samples to label from the operational context, then those samples are used as the test set to calculate the traditional accuracy estimate. Our method differs in that (1) it admits the existence of an ``outdated" test set at the beginning of the process and tackles its continuing update, and (2) it performs the accuracy prediction on the (larger) unlabeled operational data, instead of the traditional estimate from labels. 

\section{Preliminaries}
% In this section, we introduce certain concepts used in our methods and throughout the paper. 
\subsection{Sample Selection Bias Correction}
\label{sec:bias}

In the presence of sample selection bias (SSB), the empirical distribution of a finite dataset differs from its underlying population distribution. 
% \citep{zadrozny2004learning} formalized this concept in the context of machine learning 
% and outlined possible correction methods. 
Let $s$ denote a random variable indicating a particular instance is selected in the sample. 
If the probability of $s$ depends on the feature $(X)$ or label $(Y)$ variables, the selection is said to be biased. SSB is a widely studied problem \cite{elkan2001foundations, zadrozny2004learning,zadrozny2003cost, fan2005improved, dudik2006correcting} as it contributes to discrepancies between training and testing data in machine learning. A special case of interest in our investigations is the so-called 
{\em covariate shift} which assumes that the probability of $s$ only depends on the 
feature and not the label, i.e., $\Pr(s|X,Y)=\Pr(s|X)$.
\citeauthor{zadrozny2004learning} (\citeyear{zadrozny2004learning}) described an optimum bias correction method for learners 
in the presence of covariate shift via re-weighting. 

Let $D_t$ be the true distribution over $X\times Y$, and let $D_b$ be a distribution producing a 
dataset with an inherent covariate shift (selection bias with respect to $X$). 
Using the selection variable $s$, this relationship is 
expressed as: $\Pr_{D_b}(X,Y)=\Pr_{D_t}(X,Y|s=1)$. Recalling that under
covariate shift it holds that $\Pr(s=1|X,Y)=\Pr(s=1|X)$, the true distribution $D_t$ can be recovered 
from the biased distribution as follows:
\begin{subequations}
\begin{align}
    \label{eq:bias1}
        \Pr(s=1|X)&=\frac{{\Pr}_{D_t}(X,Y|s=1)\Pr(s=1)}{{\Pr}_{D_t}(X,Y)} \\
    \label{eq:bias2}
        {\Pr}_{D_t}(X,Y)&={\Pr}_{D_b}(X,Y)\frac{\Pr(s=1)}{\Pr(s=1|X)}
\end{align}
\end{subequations}
Eq. (\ref{eq:bias2}) gives the relationship between the true and biased distributions and
identifies the corrective element: a weighting factor given by
    $w_i=\frac{Pr(s=1)}{Pr(s=1|X=x_i)}\geq 0$    
which can be found for each instance $x_i$ of an unlabeled set. The quantities in the above equation need to be estimated from finite samples. An extensive work on effects of the estimation error on bias correction is given by \citeauthor{cortes2008sample} (\citeyear{cortes2008sample}).

\subsection{Calibration}
Generally speaking, the output of machine learning models can be used for diagnostic or prognostic purposes. While the only important quality for diagnostic (e.g., classification) purposes is the relative ranking of the outputs (e.g., which class attains the highest score), for prognostic purposes (e.g. accuracy estimation), the ability to output a well-calibrated quantity becomes essential. It is well known that machine learning models tend to produce more or less miscalibrated output probabilities. As a remedy, a variety of calibration techniques is available \cite{Guo2017, Zadrozny2002, bella2010calibration}.
In our case, calibration is achieved by a binning mechanism as part of the performance prediction, described in Section \ref{Sec:Performance Prediction}. 

\section{Methods}
\label{Sec:Methods}

\begin{figure}[thbp] 
\centering
\includegraphics[width=0.98\columnwidth, height=6cm]{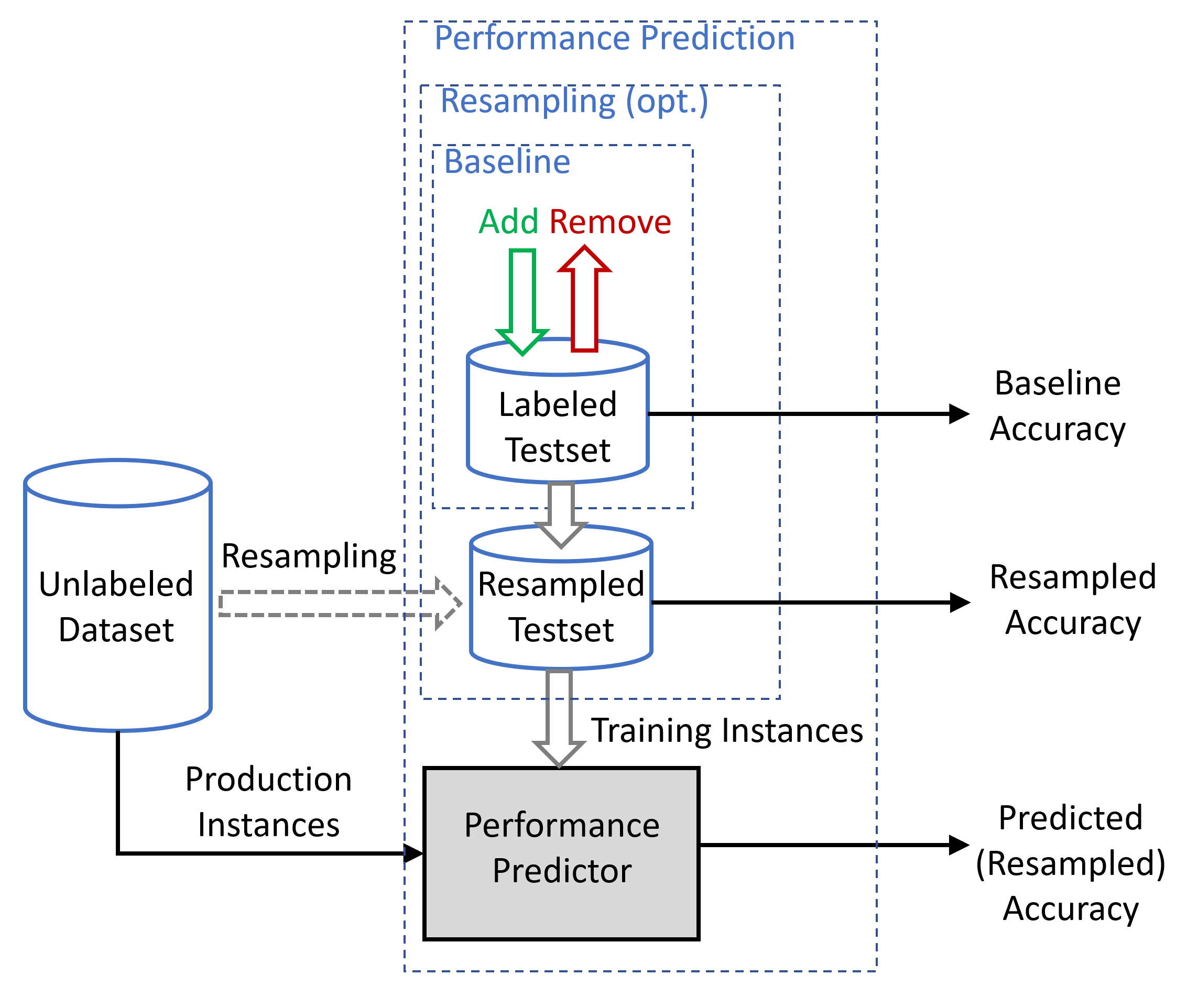} 
\caption{Overview of our approach}
\label{fig:how it works}
\end{figure}

Our overall approach to addressing the problem(s) formulated in Section \ref{Sec:Problem Formulation} is 
depicted in Figure \ref{fig:how it works}. The goal is to generate an accuracy estimate that is 
as close as possible to the true value of the entire production set. In practice, this value is unknown 
as there is no ground truth (labels) available. To tackle this, we investigate three techniques, 
separately as well as in combination: (1) Performance Prediction (described in Section \ref{Sec:Performance Prediction}), (2) Resampling (Section \ref{Sec:test set Resampling}), and (3) Prioritization of label addition and removal 
(Section \ref{Sec:Prioritization}). 
The ability of the performance predictor
to estimate the accuracy of a large amount of unlabeled data, while continually adjusting its model as new 
labels become available, plays an essential role. Making the working test set resemble 
(in feature distribution) the unlabeled production set as much as possible further improves the potential 
for an accurate estimate (see Figure \ref{fig:how it works}).

\subsection{Performance Prediction}
\label{Sec:Performance Prediction}
\begin{figure}[htbp!] 
\centering
\includegraphics[width=1.0\columnwidth]{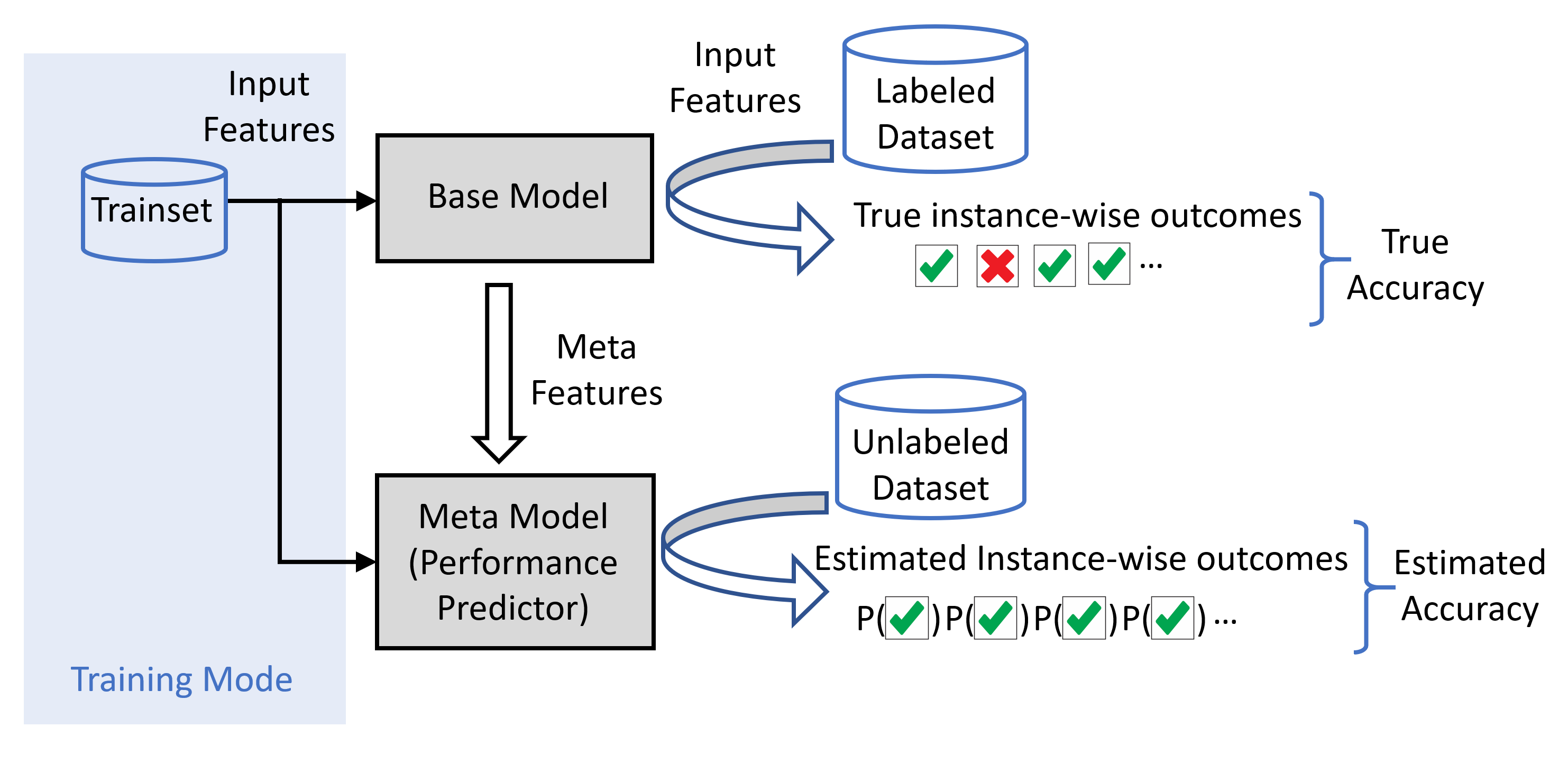} 
\caption{Performance prediction framework}
\label{fig:performance predictor}
\end{figure}

Figure \ref{fig:performance predictor} illustrates the performance predictor's functionality: Given 
a base model carrying out a classification task, the performance predictor acts 
as a meta-model observing the outputs (confidences) as well as the 
{\em outcomes} (i.e., correct vs. incorrect classifications) of the base model. 
In general, a performance predictor learns to predict the instance-wise
probability of the base model succeeding at its task, and the base model may be
considered either a whitebox or a blackbox with respect to extracting various 
meta-features (see Figure 1 of \cite{chen2019AISTATS}).
In this study, we consider a blackbox base model and adopt a
non-parametric meta-model based on binning. The training is given in Algorithm \ref{Alg:perfpred training}. Here, the confidence scores generated by the base model are assigned to 
equidistantly spaced bins. The sample mean and standard deviation of the corresponding outcomes, 
i.e., accuracy and spread, are then calculated in each bin. The totality of $M$ such bins represents
the predictor. Algorithm \ref{Alg:perfpred runtime} exercises the predictive step: Given an instance
of a confidence score, the predictor returns the mean and spread looked up in the 
corresponding bin. To obtain the cumulative (batch) estimate of accuracy, the instance-wise 
predictions are averaged over the batch.  

The above performance predictor naturally combines two functions, namely meta-learning and calibration. 
In our initial scoping experiments, this calibration based setup performed equally well as more complex parametric meta-models followed by explicit calibration. We also had experimented with other calibration techniques, for instance, the isotonic regression, which, in effect, achieves adaptive binning. Our experimental evidence showed that there were no significant differences in calibration quality (Brier score) between these methods, implying a well-behaved bin utilization of the fixed binning mechanism. In light of that we focus on the simpler of these variants to promote simplicity throughout the approach.  

\begin{algorithm}[htb]
  \caption{Train Performance Predictor}
  \label{Alg:perfpred training}
\begin{algorithmic}
 \renewcommand{\algorithmicrequire}{\textbf{Input:}}
 \renewcommand{\algorithmicensure}{\textbf{Output:}}
  \REQUIRE Instance-wise confidences: $s_1, ..., s_N$\\ Binary outcomes: $o=o_1,...,o_N$, $o_i\in\{0,1\}$ \\ Number of bins: $M$
  
  \ENSURE Predictor: $B_i=(l_i,u_i,\widehat{Acc_i}, \sigma_i)_{1\leq i \leq M}$
  
  \STATE Determine lower and upper bounds, $l_i, u_i$, for each bin.
  \FOR{sample pairs $\{s_i, o_i\}$}
  \STATE $b \leftarrow determineBin(s_i)$
  \STATE Assign $o_i$ to bin $Bin_b$
  \ENDFOR
  
  \FOR{bins $Bin_i$}
  \STATE $\widehat{Acc}_i\leftarrow Mean(o|Bin_i)$
  \STATE $\sigma_i\leftarrow Std(o|Bin_i)$
  \STATE $B_i\leftarrow(l_i, u_i, \widehat{Acc}_i, \sigma_i)$
  \ENDFOR
\end{algorithmic}
\end{algorithm}

\begin{algorithm}[htb]
  \caption{Predict Accuracy}
  \label{Alg:perfpred runtime}
\begin{algorithmic}
 \renewcommand{\algorithmicrequire}{\textbf{Input:}}
 \renewcommand{\algorithmicensure}{\textbf{Output:}}
  \REQUIRE Confidence $s$; Predictor $\{B_i\}_{1\leq i \leq M}$
  
  \ENSURE $\widehat{Acc}(s)\pm \sigma(s)$
  
  \STATE $b \leftarrow determineBin(s)$
  \STATE $\widehat{Acc}(s)\leftarrow B_b[\widehat{Acc}]$
  \STATE $\sigma(s)\leftarrow B_b[\sigma]$
\end{algorithmic}
\end{algorithm}

\subsection{Test Set Resampling}
\label{Sec:test set Resampling}
A common source of discrepancy between current production data and a test set
is covariate shift (CS). In CS, changes in the joint distribution 
of features and labels, $\Pr(X,Y)$ are explained solely by changes in $\Pr(X)$, i.e., 
$\Pr(X,Y)=\Pr(Y|X)\Pr(X)$, with $\Pr(Y|X)$ unchanged. To counter any CS present 
in our setting, 
the SSB technique described in Section \ref{sec:bias} is applied. 

Due to sample selection bias, given a labeled test set $T$ and unlabeled production set $P$, we cannot assume that $T\sim P$. 
To use Eq. (\ref{eq:bias2}) to correct for the selection bias, $\Pr(s=1)$ and $\Pr(s=1|x=x_i)$ need to be estimated. Assuming that $T\subseteq P$ \citep{cortes2008sample}, we estimate the former as $\Pr(s=1)=\frac{|T|}{|P|}$. 
To calculate $\Pr(s=1|x=x_i)$, we discretize the problem domain by using a classical technique of \textit{vector quantization} \citep{gray1984vector,soong1987report},
which maps each data vector to a representative codeword. Each of our data vectors consist of features concatenated with output class probabilities obtained from the base model's output (model confidence values).   
To construct codewords, we apply K-means clustering to the combined test and production sets, producing a set of $K$ cluster centroids (the codewords), $C=\{c_k\}_{1\leq k\leq K}$. % JN: This should go to a portion of the Experiments where we describe all details of parameter settings and how they were determined. It may not be enough to say "we picked" - we will need to say how we picked that value, to prevent reviewers picking on that:
For experiments with structured datasets, K is chosen to be equal to the size of data vector, and for image datasets, K=256 is used. 
Given a mapping from feature vectors to their respective centroids, $x_i\rightarrow c_k, c_k\in C$, the 
probability $\Pr(s=1|x_i)$ becomes $\Pr(s=1|c_k)$ and can be estimated as $\frac{t_{c_k}}{p_{c_k}}$ where $t_{c_k}$ and $p_{c_k}$ are the number of times $c_k$ is encountered in the test set and the production set, respectively. 
Finally, the weights, used as resampling ratios here, are obtained as:
\begin{equation}
    \label{eq:bias_us}
    w_i=\frac{\sfrac{|T|}{|P|}}{\sfrac{t_{c_i}}{p_{c_i}}}.
\end{equation}
Weights $w_i$ falling below a certain threshold are reset to zero. This is equivalent to deleting the corresponding (over-represented) sample from the test set. 
Based on the above weights, the resampling is performed as an upsampling procedure as follows: all 
weights, $w_i$, are divided by the smallest occuring positive weight, $w_{min}$, and rounded to closest integer to obtain the new (upsampled) count $\lfloor{\frac{w_i}{w_{min}}}\rceil$ for the $i$-th sample. 
Note, that the weight thresholding also controls the upsampled test set size.

\subsection{Prioritized Sample Addition/Removal}
\label{Sec:Prioritization}
The performance predictor trained according to Algorithm 1 on the labeled data (test set) has two outputs for each bin: the accuracy and its standard deviation. We adopt the deviation output as a proxy for the predictor's uncertainty. As such, the prediction uncertainty offers itself for label prioritization: To add, we first label samples with highest uncertainty. Doing so is expected to improve the performance predictor's quality in the 
relevant bin (this is addressed empirically in Section \ref{sec:results}). Similarly, to delete an element
of the current test set, we prioritize samples with the lowest uncertainty values.

\begin{table*}[tb!]
\normalsize
  \centering
  \caption{Experimental Design: Data Sets}
  \resizebox{\textwidth}{!}{
    \begin{tabular}{M{4em} M{10em} M{4.5em} M{3em} M{3em} M{3em} M{3em} M{7.5em}}
    \Xhline{1pt}
    \multicolumn{1}{M{5em}}{\textbf{Data Set}} & \multicolumn{1}{M{10em}}{\centering{\textbf{Splitting Condition}}} & 
    \multicolumn{1}{M{4.5em}}{\centering\textbf{\# Features}} & \multicolumn{1}{M{3em}}{\centering\textbf{Test Size}} & \multicolumn{1}{M{3em}}{\centering\textbf{Prod Size}} &  \multicolumn{1}{M{3em}}{\centering\textbf{Train Size}} & \multicolumn{1}{M{3em}}{\centering\textbf{\# classes}} & {\textbf{Classifier}} \\
    
    \Xhline{1pt}
    \multicolumn{1}{M{9em}}{\textbf{Fashion MNIST}} & Class $\in \{0, 1, 2, 3, 4\}$ & 784 & 1403  & 4210  & 27786 & 10    & LeNet \\
\hline
    \multicolumn{1}{M{11em}}{\textbf{Credit Card Default}} & 1st Month Status = 0 & 23  & 590   & 1770    & 11682 & 2     & Random Forest \\
    \hline
    \multicolumn{1}{M{9em}}{\textbf{Bank Marketing }}& \# of employees $> 5150$ & 20 & 1033  & 3100    & 20460 & 2     & Random Forest \\
    \hline
       \multicolumn{1}{M{9em}}{\textbf{Bank Marketing }}  & Contact = ``Cell" & 20 & 553 & 1660  & 10956 & 2     & Random Forest \\
    \Xhline{1pt}
    \end{tabular}%
 }
  \label{tab:data}%
\end{table*}%

\begin{figure}[t!] 
\centering
\includegraphics[width=0.98\columnwidth]{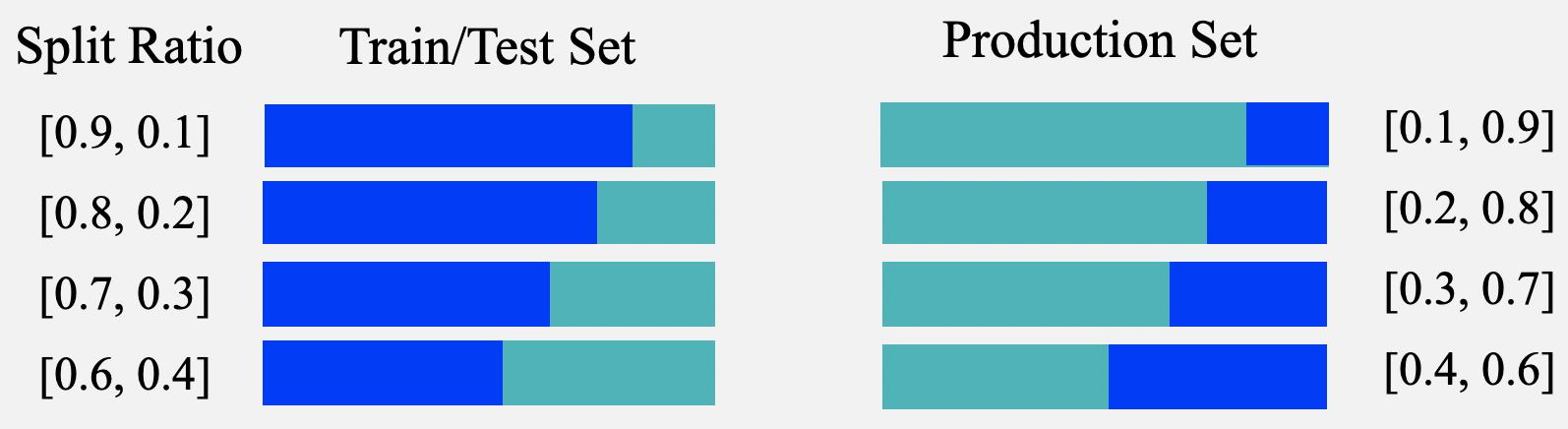} 
\caption{Dataset splits based on a biasing feature. Symmetries of those cases are included for experiments as well, making 8 total biased split pairs.}
\label{fig:bias split}
\end{figure}
\section{Experimental Design}

\subsection{Datasets and Base Models}

To study the methods described in Section \ref{Sec:Methods} on a variety of tasks and data types, we adopted three datasets: (1) {\bf Bank Marketing}\footnote{https://www.kaggle.com/janiobachmann/bank-marketing-dataset}, 
(2) {\bf Default of Credit Card Clients}\footnote{https://www.kaggle.com/uciml/default-of-credit-card-clients-dataset}, and
(3) {\bf Fashion MNIST}\footnote{https://www.kaggle.com/zalando-research/fashionmnist}.
Table \ref{tab:data} summarizes the dataset statistics, type of base model, and splitting condition (explained in Section \ref{Sec:Biased Splitting}).
As base models (i.e., models performing the base task), we trained a LeNet model \citep{lecun1998gradient} for the image classification task (Fashion MNIST), and random forest models \citep{breiman2001random} for the others. 

The LeNet model was trained for 50 epochs with a batch size of 128 and a SGD optimizer with learning rate 0.002 and momentum 0.5. The random forest models used 100 trees with a max depth of 2. 

\subsection{Biased Splitting}
\label{Sec:Biased Splitting}
Traditional test sets collected during a one-time model validation process are ineffective if they differ significantly from incoming production data. To validate our methods with respect to distributional differences between 
the test and production sets, we employ a notion of a ``biasing feature" to partition datasets into train, test, and production sets as follows: 
For a chosen biasing feature, a binary criterion is set up (``Splitting Condition" in Table \ref{tab:data}), which is used to partition the data into bins $A$ and $B$. We then proceed to construct train, test, and production sets by sampling a range of different proportions from these two bins. 
Specifically, the train and test sets are both sampled $k\%$ from bin $A$ and $(100-k)\%$ from bin $B$, while the production set is given by the opposite, $(100-k)\%$ from bin $A$ and $k\%$ from bin $B$, for $k \in \{10, 20, 30, 40, 60, 70, 80, 90\}$. 
Figure \ref{fig:bias split} illustrates this splitting logic for one half of the symmetry. Each such split represents a practical scenario where the base model's training conditions differ from the operational conditions in the production set. Our experiments are carried out for each split independently.

\subsection{Test Set Labeling and Reporting}
\label{Sec:test set labeling and reporting}
After obtaining the biased splits, we carve out a uniformly random subset of the production set with size equal to $\vert test\vert$ as a designated ``pool" to be labeled (referred to as \textit{pool set}). 
New candidates from the pool set are drawn in batches, are ``labeled," i.e. their labels are revealed, and are added to the test set. The complement of the pool set is fixed and serves as a representative of the operational domain 
whose accuracy we aim to approximate, as defined in Eq. (\ref{Eq:AccDiff}). We continue to refer to this 
complement as the production set. 

As the main experimental outcomes, we report the absolute difference between the trained base model's actual accuracy,
measured on the production set, i.e., using its all labels, and the output of the performance predictor which 
uses only the current test set as a labeled source and the production set as an unlabeled source. 
A contrastive comparison is made to a traditional approach taking into account the test set only (baseline). 
All experiments are repeated four times and averages and their spreads are reported to account for noise due to the  random selection involved in data partitioning.

\subsubsection{Add-only} 

To mimic the practical case of continuous test set update using small samples of production data 
 at a time (after manual labeling) in our experiments, we start with the initial biased test set and gradually expand it 
using samples from the pool set. This expansion is performed in 40 iterations (minibatches) with each iteration 
having a labeling budget of $\frac{|pool|}{40}$ labels. 
Hence, with a sufficiently large pool set, the test set and the production distributions will converge. 
In a first variant of ``Add-only," the minibatches are selected at random. In a second variant, 
a minibatch is formed by prioritizing samples with highest uncertainty as generated by the 
performance predictor (see Section \ref{Sec:Prioritization}). 
Since we set $|pool|=|test|$, at the end of 40 iterations our initial test set, which is biased by construction, still makes up half of all samples. 

\subsubsection{Add-delete} 
To allow for a complete domain alignment between the test set and the production set at the end of the iterative process, we consider another updating variant including sample deletion.
In this ``Add-delete" variant, each iteration adds a minibatch from the pool set to the test set, and removes a minibatch of equal size from the original test set.  
Similar to the above, we investigate two strategies for addition and deletion: random selection and a prioritized selection. In the case of deletion, samples with the smallest performance prediction uncertainty are removed first (with the intuition that such samples may be over-represented in the test set). 
Since $|pool|=|test|$, at the end of 40 iterations, all of the samples in the initial test set will be deleted and we expect both the baseline system and the performance predictor to converge to the true accuracy of the trained model on the production set.

\section{Results}
\label{sec:results}
Given the four biased dataset splits (as per Table \ref{tab:data}) exercising eight different bias proportions (see Figure \ref{fig:bias split}),  
we conduct experiments for the four labeling strategies as follows: (1) add-only with random selection, 
(2) add-only with prioritization, (3) add-delete with random selection, and (4) add-delete with prioritization, yielding a total of 128 experiments. 
The performance predictor, trained using the (labeled and gradually changing) test set, makes instance-wise predictions on all production samples with subsequent averaging to obtain the accuracy estimate,
$\widehat{Acc}$ in Eq. (\ref{Eq:AccDiff}).
In addition to the performance prediction and prioritization, the resampling method described
in Section \ref{Sec:test set Resampling} is optionally applied. 
The corresponding quantization codebook is calculated at the first iteration and kept fixed from there on. 
With resampling active, the test set is resampled according to the weights calculated according to Eq. (\ref{eq:bias_us}) at each iteration. 
Outcomes of the various experiments are presented on charts showing the absolute difference 
between predicted and actual accuracy (y-axis) as a function of points labeled 
at a given iteration (x-axis).  
In these charts, a curve shows the average value of four repetitions of the same experiment 
(see Section \ref{Sec:test set labeling and reporting}) and their
shaded bands reflect the standard deviation over the same repetitions. 
\begin{figure*}[t!]
        \centering
        \begin{subfigure}[b]{0.475\textwidth}
            \centering
            \includegraphics[width=\textwidth]{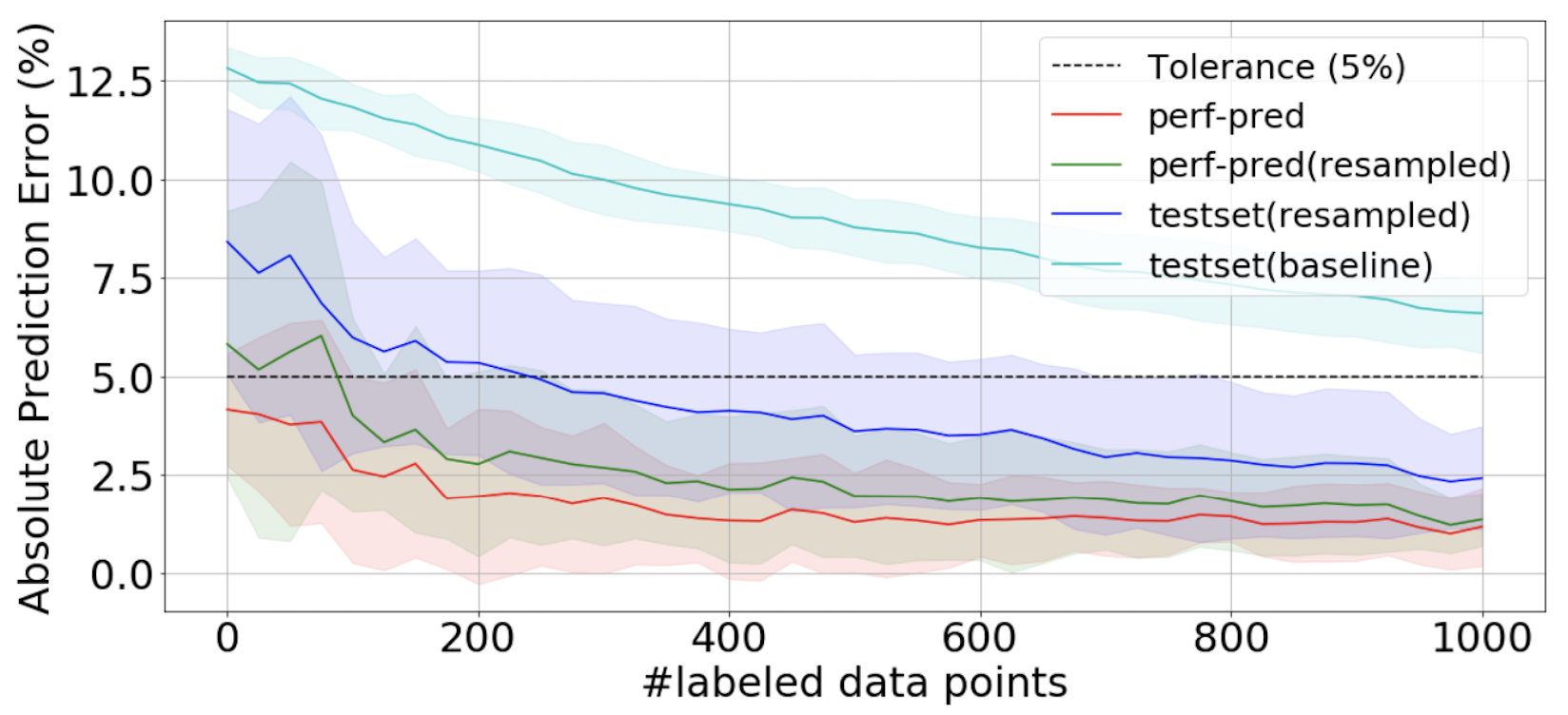}
            \caption[]%
            {{\small Add-only labeling \& Random batch selection, 10\%-90\% split}}    
            \label{fig:ao_rand}
        \end{subfigure}
        \quad
        \begin{subfigure}[b]{0.475\textwidth}  
            \centering 
            \includegraphics[width=\textwidth]{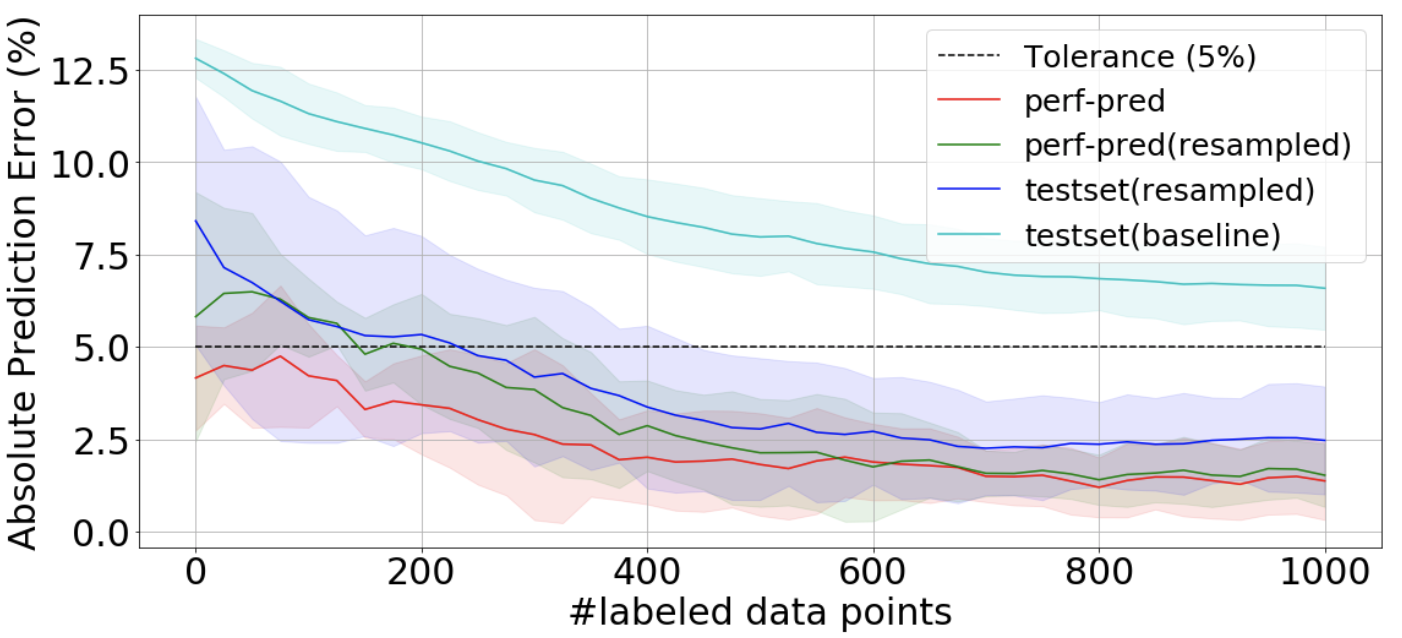}
            \caption[]%
            {{\small Add-only labeling \& Prioritized batch selection, 10\%-90\% }}    
            \label{fig:ao_sel}
        \end{subfigure}
        \vskip\baselineskip
        \begin{subfigure}[b]{0.475\textwidth}  
            \centering 
            \includegraphics[width=\textwidth]{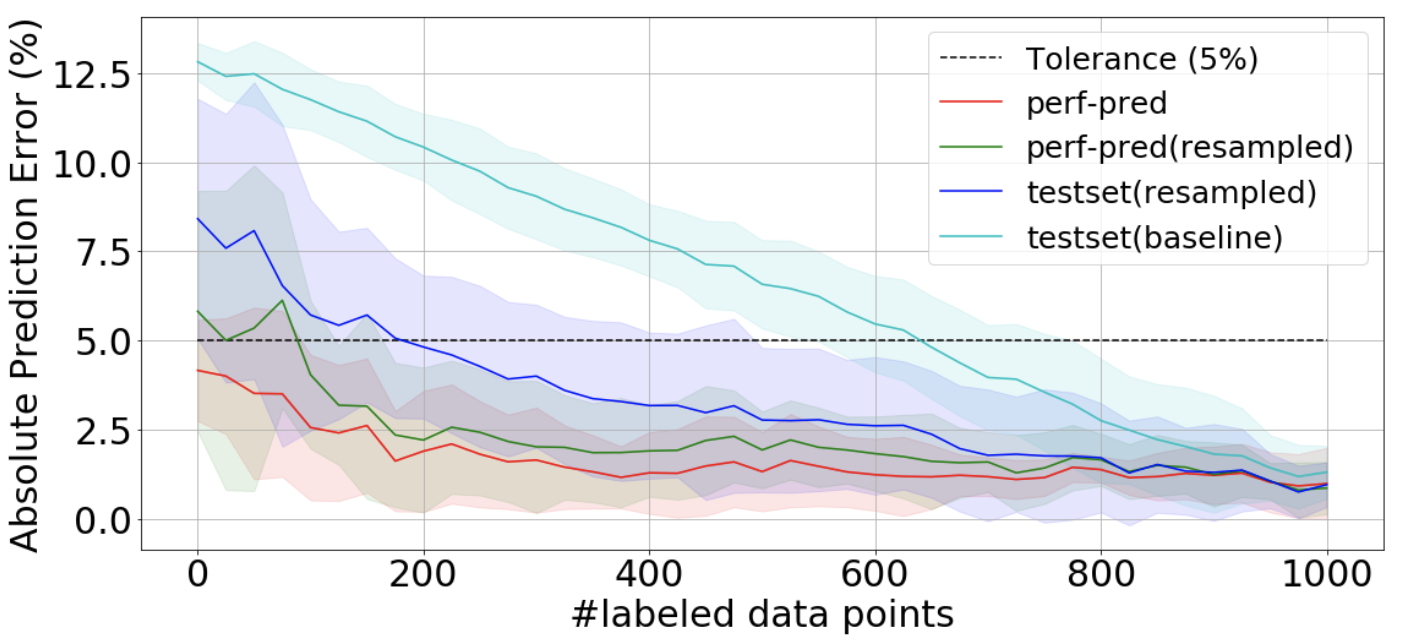}
            \caption[]%
            {{\small Add-delete labeling \& Random batch selection, 10\%-90\% }}    
            \label{fig:io_rand}
        \end{subfigure}
        \quad
        \begin{subfigure}[b]{0.475\textwidth}  
            \centering 
            \includegraphics[width=\textwidth]{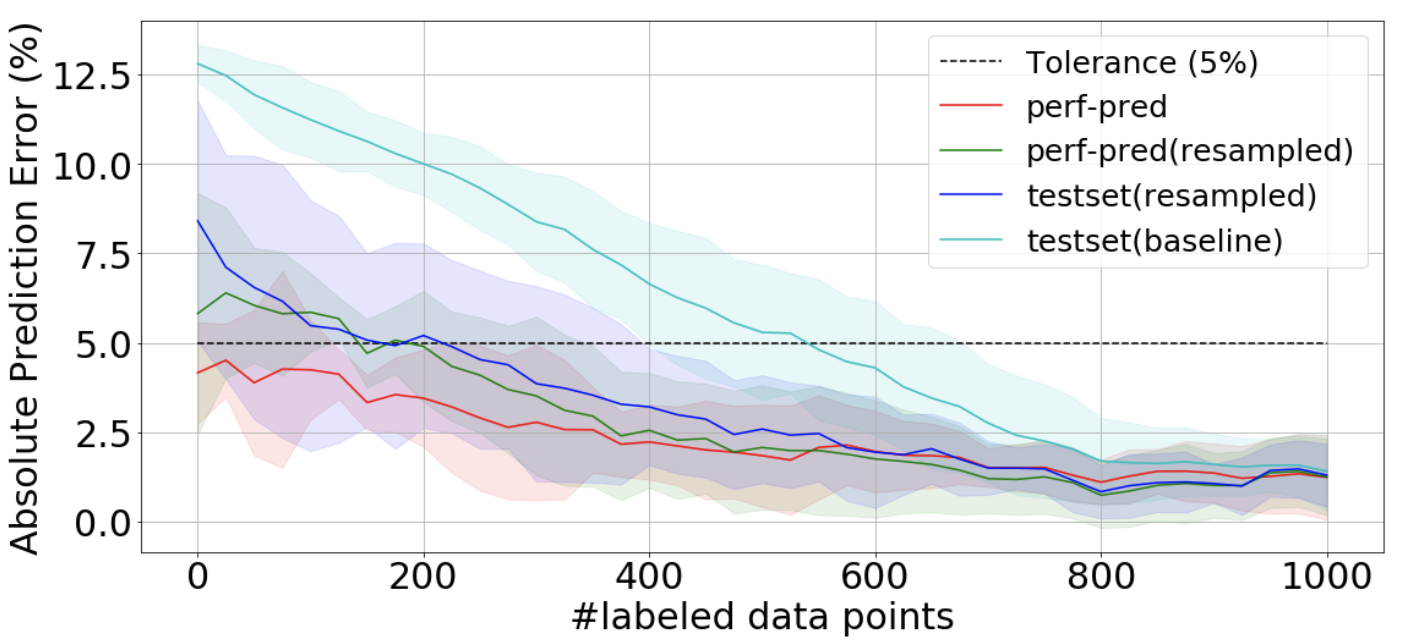}
            \caption[]%
            {{\small Add-delete labeling \& Prioritized batch selection, 10\%-90\% }}    
            \label{fig:io_sel}
        \end{subfigure}
        \vskip\baselineskip
        \begin{subfigure}[b]{0.475\textwidth}
            \centering
            \includegraphics[width=\textwidth]{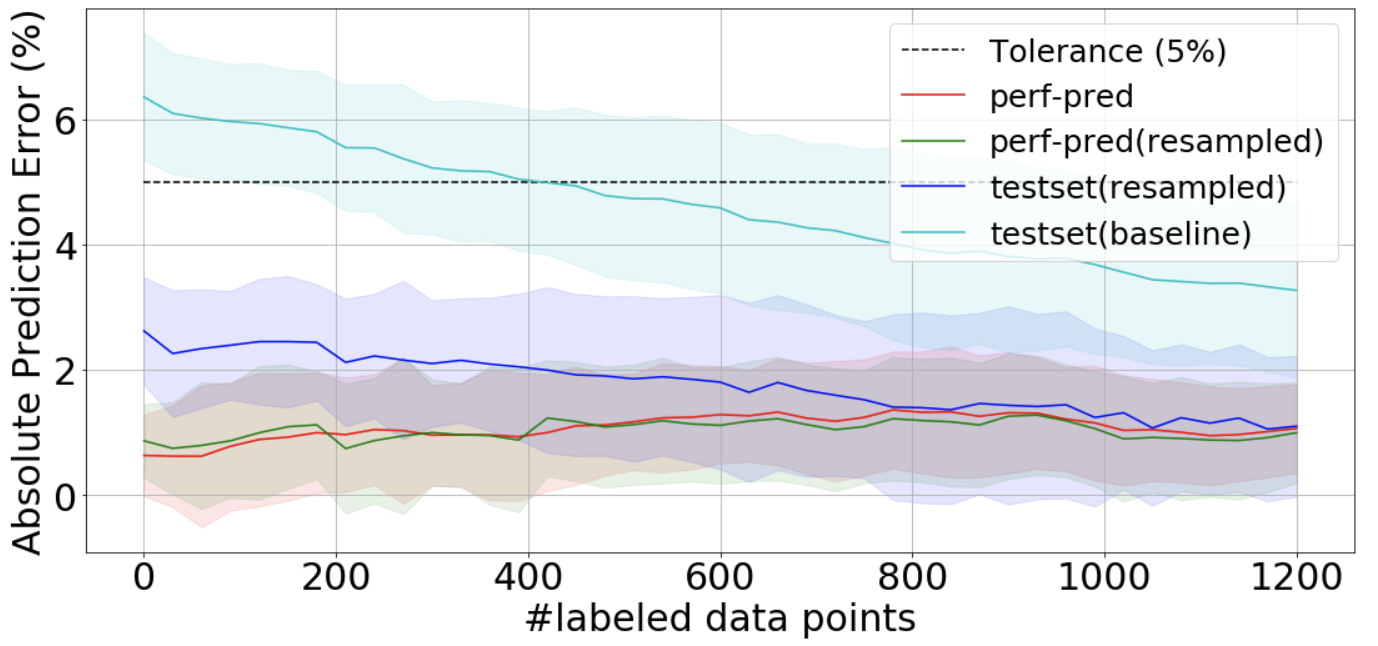}
            \caption[]%
            {{\small Add-only labeling \& Random batch selection, 30\%-70\% split}}    
             \label{fig:ao_rand2}
        \end{subfigure}
        \quad
        \begin{subfigure}[b]{0.475\textwidth}  
            \centering 
            \includegraphics[width=\textwidth]{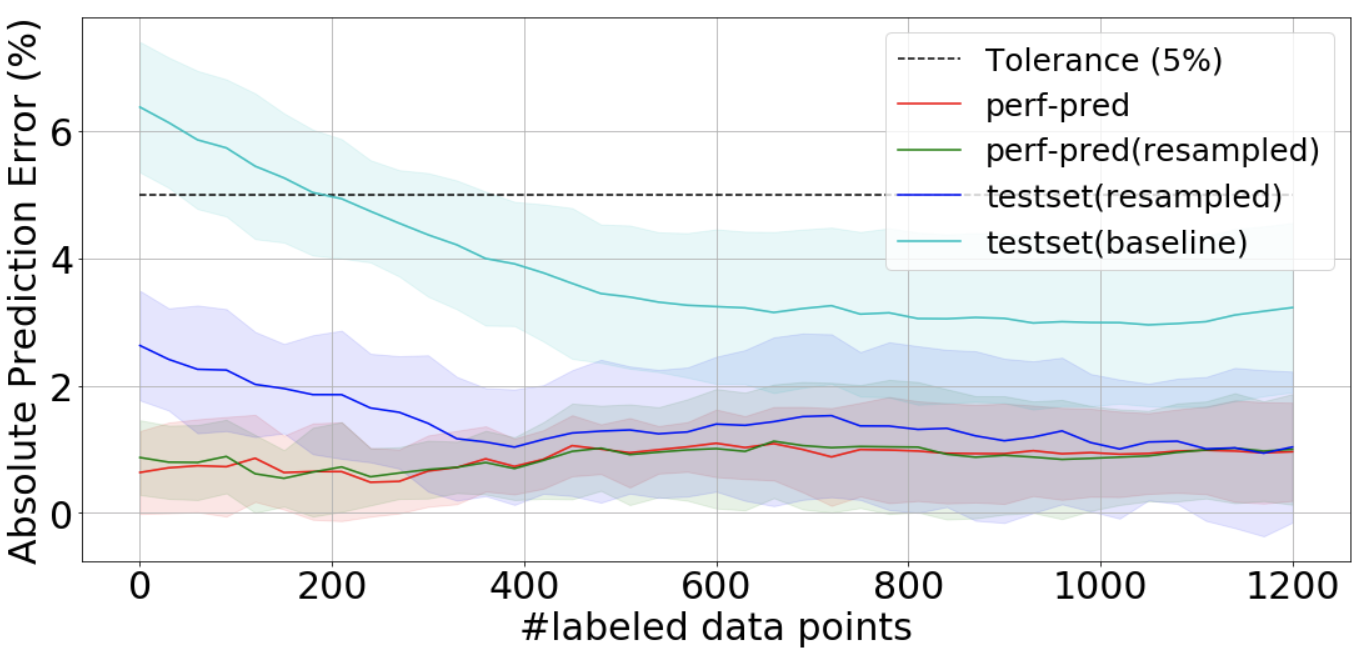}
            \caption[]%
            {{\small Add-only labeling \& Prioritized batch selection, 30\%-70\% }}    
             \label{fig:ao_sel2}
        \end{subfigure}
        \vskip\baselineskip
        \begin{subfigure}[b]{0.475\textwidth}  
            \centering 
            \includegraphics[width=\textwidth]{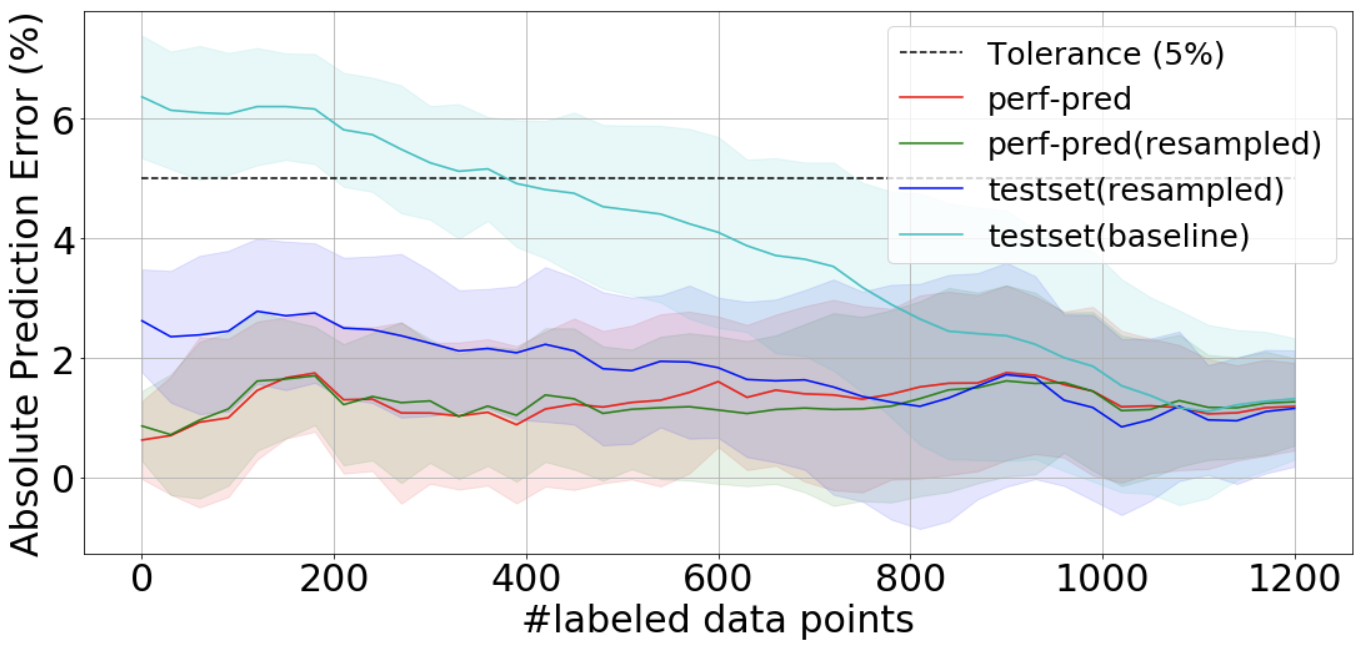}
            \caption[]%
            {{\small Add-delete labeling \& Random batch selection, 30\%-70\% }}    
            \label{fig:io_rand2}
        \end{subfigure}
        \quad
        \begin{subfigure}[b]{0.475\textwidth}  
            \centering 
            \includegraphics[width=\textwidth]{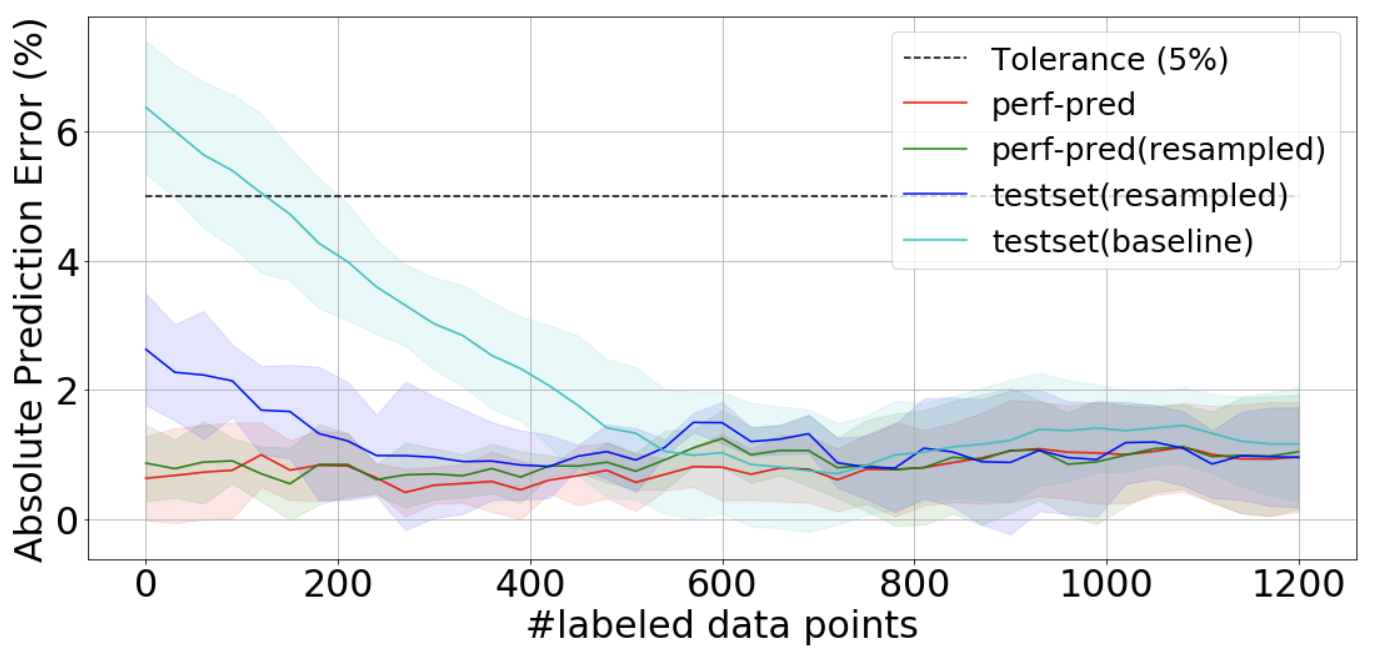}
            \caption[]%
            {{\small Add-delete labeling \& Prioritized batch selection, 30\%-70\% }}    
             \label{fig:io_sel22}
        \end{subfigure}
        \caption{ \centering Number of newly labeled samples (x-axis) vs. Accuracy prediction error (y-axis) \newline Dataset: Bank Marketing, \# of employees based split. (a-d): 10\%-90\% biased split 
        (e-h): 30\%-70\%.}
        \label{fig:results}
\end{figure*} 

 As a representative example, Figure \ref{fig:results} shows 
 charts for the Bank Marketing dataset with test set bias proportions of 30\%-70\% (moderate split) and 10\%-90\% (extreme split), and the four labeling strategies (add-only, add-only prioritized, add-delete, add-delete prioritized). 
 Each chart contains four curves corresponding to the proposed algorithms and 
 the baseline, as follows: (1) Performance predictor without resampling (``perf-pred"), 
 (2) Performance predictor with resampling (``perf-pred (resampled)"), (3) Accuracy on the test set (``test set"), 
 and (4) Accuracy on a resampled test set (``test set (resampled)"). 
 As can be seen in all charts, the two methods utilizing the performance predictor, 
 namely ``perf-pred (resampled)" and ``perf-pred", 
 consistently attain the lowest accuracy error as the number of labels grows. Resampling alone 
 (``test set (resampled)") also brings a considerable improvement over the baseline (conventional 
 test accuracy). Furthermore, all four charts show a clear trend of convergence 
 between the actual and predicted accuracy, as expected. Among the four, the ``Add-Delete" strategy and, in particular, 
 the ``Add-Delete with Prioritization" strategy dominate. 
Note that the baseline method in Figures \ref{fig:ao_sel}, \ref{fig:io_sel}, \ref{fig:ao_sel2}, \ref{fig:io_sel22} correspond to a classical accuracy 
calculation except the test set updates follow the uncertainty-based prioritization generated by the performance predictor.
\begin{figure}[t] 
	\centering
	\includegraphics[width=0.98\columnwidth]{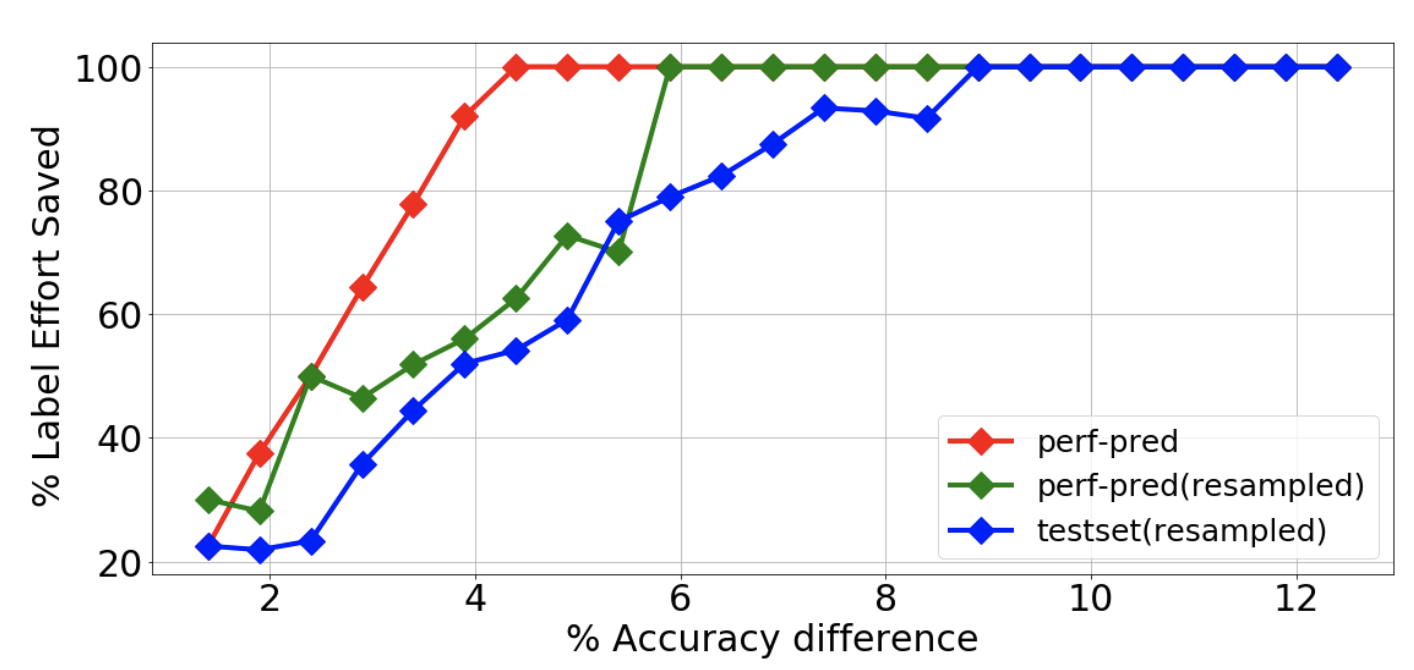} 
	\caption{\% Labeling effort saved as a function of desired accuracy tolerance for the example in Fig. \ref{fig:io_sel}}
	\label{fig:io_sel2}
\end{figure}

\begin{table*} [h!]
	\centering
	\caption{Results Overview: Area Under Curve (AUC) and rank order statistics (lower values are desirable) over all datasets and splits. Values$^{*}$ indicate difference at $p<0.001$ to the respective best value. Ranks reflect instance-wise ordering of the methods (rows) by their AUC. }
	\resizebox{\textwidth}{!}{
	\begin{tabular}{lccccc}
	\Xhline{1pt}
    \textbf{Predict Method} & \multicolumn{1}{M{6.75em}}{\textbf{add-delete (prioritized)}} & \multicolumn{1}{M{5.75em}}{\textbf{add-delete (random)}} & \multicolumn{1}{M{6.5em}}{\textbf{add-only (prioritized)}} & \multicolumn{1}{M{6.3em}}{\textbf{add-only (random)}} & \multicolumn{1}{M{6.3em}}{\textbf{Overall}}\\
    \Xhline{1pt}
	&\multicolumn{4}{c}{ Average AUC / Row-Rank Order} &\\
    \hline
    perf-pred (resampled)           & {\bf 83.3} / {\bf 1.7} & {\bf 79.3} / 2.0 & {\bf 93.3} / {\bf 1.8}  & {\bf 90.8} / {2.0} & {\bf 86.7} / {\bf 1.8}\\
    perf-pred                      & 96.1 / 2.2 & 93.8 / {\bf 1.9} & 113.4 / 2.1 & 112.6 / {\bf 1.9} & 104.0 / 2.0 \\
    traditional testing (resampled) & 92.6 / 2.5 & 99.4$^{*}$ / 2.3$^{*}$ & 113.5$^{*}$ / 2.5$^{*}$ & 119.4$^{*}$ / 2.4$^{*}$ & 106.2$^{*}$ / 2.4$^{*}$\\
    traditional testing (baseline) & 173.0$^{*}$/ 3.7$^{*}$ & 167.9$^{*}$ /3.7$^{*}$ & 208.6$^{*}$ / 3.7$^{*}$ & 219.4$^{*}$ / 3.8$^{*}$ & 192.2$^{*}$ / 3.7$^{*}$\\
    \hline
    Overall                 & 111.2 & {\bf 110.1} & 132.2$^{*}$ & 135.6$^{*}$\\    
    \Xhline{1pt}
    \end{tabular}%
    }
	\label{tab:all results}%
\end{table*}

\begin{figure}[t] 
	\centering
	\includegraphics[width=0.98\columnwidth]{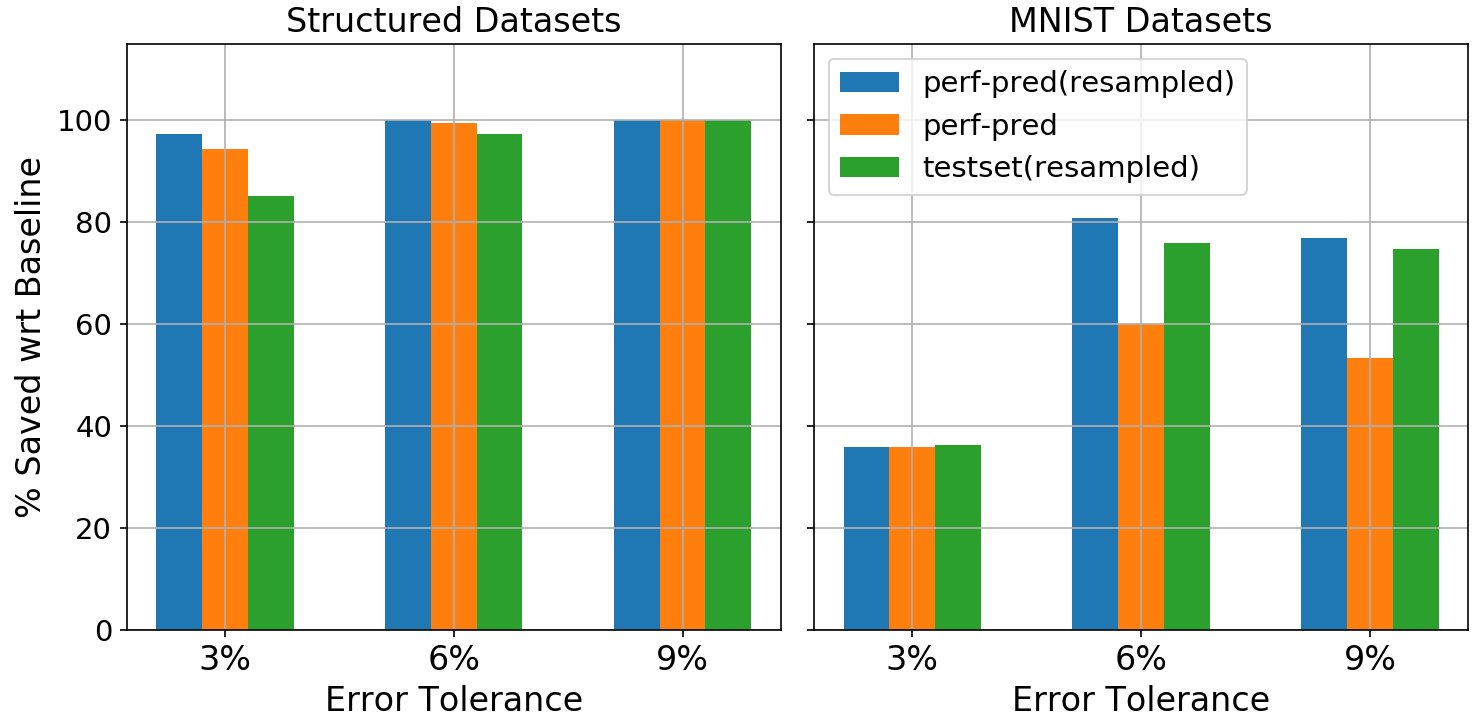} 
	\caption{Average labeling effort saved with respect to the baseline for add-delete labeling \& random batch selection}
	\label{fig:overall_effort_saved}
\end{figure}

\begin{table}[htbp!]
	\centering
	\caption{Effect of test set vs. Production mismatch (AUC and ranks averaged over test sets and update strategies). The extreme subset includes
	20\%-80\% and 10\%-90\%, the moderate subset includes 70\%-30\%, 60\%-40\% proportions (with their symmetric counterparts). Values$^{*/\dagger}$ indicate difference at $p<0.001/p<0.05$ to the respective best value. }
	\resizebox{\columnwidth}{!}{
	\begin{tabular}{lcc}
	\Xhline{1pt}
    \textbf{Predict Method} & \multicolumn{1}{M{4.75em}}{\textbf{Moderate Subset}} & \multicolumn{1}{M{4.75em}}{\textbf{Extreme Subset}}\\
    \Xhline{1pt}
	&\multicolumn{2}{c}{ Average AUC / Row-Rank Order} \\
    \hline
     {perf-pred (resampled) }          & 50.1 / {\bf 1.9} & {\bf 123.3} / {\bf 1.8}\\
    perf-pred                      & {\bf 48.5} / 2.0  & 159.4$^\dagger$ / 2.0\\
    {traditional testing (resampled)} & 62.3$^{*}$ / 2.4$^{*}$ & 150.1$^{*}$ / 2.4$^{*}$ \\
    {traditional testing (baseline)} & 101.1$^{*}$ / 3.7$^{*}$ & 283.2$^{*}$ / 3.8$^{*}$ \\
    \Xhline{1pt}
    \end{tabular}}
	\label{tab:imbalance results}
\end{table}
 The results in the Figure \ref{fig:results} also show the aspect of saving labeling effort: suppose
 we are given a tolerance band for the accuracy estimate of, say, 5\% (horizontal line in the charts), 
 representing the maximum acceptable discrepancy in accuracy. 
 On the Bank Marketing Data, we can conclude that to satisfy the tolerance criterion of 5\% we 
 save approximately 150-650 labeling operations, depending on the specific strategy and bias ratio. 
 Another way of organizing the information shown in Figure \ref{fig:results} is expressing the potential 
{\em labeling effort saved} as a function of the accuracy difference. This trade-off is shown in 
Figure \ref{fig:io_sel2} for the same example as in the Figure \ref{fig:io_sel}. 
This chart provides a user with the information about the proportion of labeling operations each of the 
three techniques would save (y-axis), had the user's accuracy tolerance threshold been a certain 
value (x-axis). The relative saving is calculated with respect to the baseline (``test set") accuracy
as: $100\cdot(N^{test set}(Acc)-N^m(Acc))/N^{test set}(Acc)$ where $N^{test set}(Acc)$ denotes the number of labels needed for the ``testset" 
baseline to reach accuracy of $Acc$, and $N^{m}(Acc)$ is same 
quantity but for the contrasted technique, i.e., $m=$\{perf-pred, perf-pred(resampled), test set(resampled)\}.
Note that since perf-pred methods achieve less than 6\% prediction error without any labels added, labeling effort saved is 100\% for $Acc>6\%$. 
  To allow for more comprehensive conclusions we summarize each curve (experiment) using the 
 {\em area under the curve} (AUC). This metric calculates the area under each curve on the coordinates
 exemplified in Figure \ref{fig:results}, namely the labeling effort (x-axis) and accuracy error (y-axis), and is not to be confused with an AUC of a receiver operating curve.
 Since both axes correspond to a cost, small AUC values are desirable. 

 Results in terms of the AUC metric are summarized in Table \ref{tab:all results}. The values in this
 table are averages over all proportion splits ranging from 90\%-10\% to 10\%-90\% showing the effectiveness
 of each method (rows) and update strategy (columns).  
 Since the AUC magnitude varies with each experiment due to the hardness of the individual tasks, 
 we also compute the rank order statistics for each of the four methods (rows in Table \ref{tab:all results}) per 
 experiment, i.e., best (worst) performing method per instance receives rank 1 (4), and report its average along with the AUC. 
 Averages over rows and columns are also shown. Values with an asterisk are statistically significant ($p<0.001$) from the best value in the corresponding column. Statistical significance was determined using the Wilcoxon signed-rank test.
Several conclusions can be made from Table \ref{tab:all results}: (1) The performance predictor with resampling produces the 
best AUC averages overall (AUC=86.7), albeit the next alternative without resampling (AUC=104.0) seems to be not significantly different. 
(2) Resampling alone can improve over the traditional testing significantly, reducing the AUC from 192.2 to 106.2. 
(3) The ``add-delete" with and without prioritization are a statistical tie and are significantly different from their ``add-only"
variants, indicating data removal plays an important role in an effective test set maintenance. 

Figure \ref{fig:overall_effort_saved} shows the overall labeling effort saved due to the individual methods, 
similar to the Figure \ref{fig:io_sel2} above, for all splits of the structured datasets, and all splits of
the F-MNIST dataset. Due to each dataset and split giving rise to a different range of achievable accuracies, 
a binning to three values, namely up to 3\%, 6\%, and 9\%, was applied. The F-MNIST dataset is plotted 
separately as its underlying accuracy range differs significantly from the structured datasets. 
As seen in Figure \ref{fig:overall_effort_saved}, the savings depend on the choice of tolerance
and are high in the case of structured datasets (at 100\% for 6+\% tolerance), and somewhat lower 
on F-MNIST (80\% saving at 6\% tolerance). In the F-MNIST case, it also appears that resampling 
plays a more important role. We believe this to be a consequence of the F-MNIST splitting criterion 
(see Table \ref{tab:data}) directly tied to the target classes, thus producing imbalance more amenable 
to bias correction. 

Another aspect of interest is the degree of imbalance in the proportion splits. To study this aspect, we group the splits
into a "Moderate" (splits 30\%-70\%, 40\%-60\%, 60\%-40\%, 70\%-30\%) and an "Extreme" (10\%-90\%, 20\%-80\%, 
80\%-20\%, 90\%-10\%) group. Table \ref{tab:imbalance results} shows the AUC and rank averages over all datasets and 
update strategies, with respect to the two groups. Superscripts $*/\dagger$ denote statistical significance at the 
$p<0.001$/ $p<0.05$ levels determined via the Wilcoxon signed-rank test, as above. 
While these results confirm the relative ranking of the methods 
observed in Table \ref{tab:all results}, they also suggest that, in the more Extreme cases, the benefit of 
resampling gains importance, in particular in conjunction 
with the performance predictor (AUC=123.3). The predictor without resampling receives a significantly larger AUC of 159.4 - a very different result compared to the Moderate case. A second observation relates to the magnitude of the AUC values: the Extreme subset 
induces AUC values that are multiples of those observed in the Moderate group, reflecting the hardness of the underlying 
task in more extreme mismatch scenarios. It is reassuring to observe that the relative ranking of the individual methods
remains stable across the two conditions.

\section{Conclusions}

Despite their ubiquitous use, test sets are a poor mechanism for evaluating model quality in real world applications.
Without constant monitoring and updating, test sets can drift from production traffic, making the resulting test results irrelevant and misleading.
Our work directly addresses this problem by proposing a set of techniques (resampling, performance prediction, and prioritized addition/removal) which drastically reduce the effort required to achieve a given level of testing quality.
On multiple datasets of different modalities, these techniques resulted in a labeling effort reduction ranging between 80\% and 100\%, depending on dataset, for an error tolerance of 5\%.%

These results were achieved using a simple confidence-binning-based performance predictor. Even without performance prediction, simply resampling the test set to match production data and employing an ``add and remove'' strategy showed significant improvements over a traditional test set approach.
We believe this result is significant because it offers strong evidence for practitioners to move away from the time honored tradition of test sets, instead embracing performance predictors to estimate their model quality.  It also encourages researchers to innovate further in this space, identifying new techniques for measuring model quality more accurately and with lower cost.

\bibliography{bibliography}
\bibliographystyle{icml2020}

\end{document}